\documentclass[runningheads]{llncs}

\usepackage{accv}

\usepackage{accvabbrv}

\usepackage{graphicx}
\usepackage{booktabs}

\usepackage{indentfirst}
\usepackage{multirow}
\usepackage{makecell}

\usepackage[accsupp]{axessibility}  % Improves PDF readability for those with disabilities.

\usepackage{hyperref}

\usepackage{orcidlink}

\begin{document}

% ---------------------------------------------------------------
% TODO REVIEW: Replace with your title
% \title{Author Guidelines for ACCV Submission} 
\title{QCLR: A Plug-in Robustification Module via Quantization for Contrastive Learning in Self-Supervised Skeleton-Based Human Action Recognition}

% TODO REVIEW: If the paper title is too long for the running head, you can set
% an abbreviated paper title here. If not, comment out.
% \titlerunning{Abbreviated paper title}
\titlerunning{QCLR}

% TODO FINAL: Replace with your author list. 
% Include the authors' OCRID for the camera-ready version, if at all possible.
% \author{First Author\inst{1}\orcidlink{0000-1111-2222-3333} \and
% Second Author\inst{2,3}\orcidlink{1111-2222-3333-4444} \and
% Third Author\inst{3}\orcidlink{2222--3333-4444-5555}}
\author{Nanzhu Chen\inst{1}\orcidlink{0009-0000-0002-7359} \and
Feixiang He\inst{2}\orcidlink{0009-0000-9475-6834} \and
Yunfeng Diao\inst{3}\orcidlink{0000-0002-9455-1510} \and
Duygu Sarikaya\inst{1}\orcidlink{0000-0002-2083-4999} \and
He Wang\inst{4}\orcidlink{0000-0002-2281-5679}}

% TODO FINAL: Replace with an abbreviated list of authors.
% \authorrunning{F.~Author et al.}
\authorrunning{N.~Chen et al.}
% First names are abbreviated in the running head.
% If there are more than two authors, 'et al.' is used.

% TODO FINAL: Replace with your institution list.
% \institute{Princeton University, Princeton NJ 08544, USA \and
% Springer Heidelberg, Tiergartenstr.~17, 69121 Heidelberg, Germany
% \email{lncs@springer.com}\\
% \url{http://www.springer.com/gp/computer-science/lncs} \and
% ABC Institute, Rupert-Karls-University Heidelberg, Heidelberg, Germany\\
% \email{\{abc,lncs\}@uni-heidelberg.de}}
\institute{University of Leeds, Leeds, United Kingdom\\
\email{\{scnch,d.sarikaya\}@leeds.ac.uk} \and
Central South University, Changsha, China\\
\email{feixiang.he@csu.edu.cn} \and
Hefei University of Technology, Hefei, China\\
\email{diaoyunfeng@hfut.edu.cn} \and
University College London, London, United Kingdom\\
\email{he\_wang@ucl.ac.uk}}

\maketitle

\begin{abstract}
  % The abstract should concisely summarize the contents of the paper. 
  % While there is no fixed length restriction for the abstract, it is recommended to limit your abstract to approximately 150 words.
  % Please include keywords as in the example below. 
  % This is required for papers in LNCS proceedings.
  Skeleton-based human action recognition has gained increasing attention due to the stability of skeleton data across diverse environments and advances in motion capture technologies. Following the success of supervised learning approaches, self-supervised learning approaches, particularly contrastive learning ones, have achieved impressive performance. However, they remain vulnerable to adversarial attacks. In this work, we propose a plug-in robustification module via quantization for contrastive learning in self-supervised skeleton-based human action recognition (QCLR), which enhances the robustness of representations against adversarial attacks while preserving their representational capability. Specifically, we employ a dual-level quantization scheme across the encoding pipeline, which breaks the gradient propagation and absorbs small adversarial perturbations within the quantization bins. To mitigate the impact of inevitable quantization error, we further design an attention-guided multi-faceted quantization strategy to maintain the representation quality. Experiments on several benchmark datasets demonstrate that integrating QCLR into a progressive lineage of mainstream contrastive learning models substantially improves robustness against both gradient-based white-box and gradient-free black-box adversarial attacks, while preserving highly competitive classification accuracy.
  % \keywords{First keyword \and Second keyword \and Third keyword}
\end{abstract}

\section{Introduction}

In recent years, human action recognition has become an active topic of research in computer vision due to its wide range of applications \cite{wang2013dense, luo2013group, hu2015jointly}. Skeleton-based human action recognition, in particular, has gained considerable attention due to the stability of skeleton data in different environments and the advances in motion capture technologies that enhance fidelity \cite{liu2016spatio, wang2017modeling, soo2017interpretable, yan2018spatial, shi2019two, liu2020disentangling, chen2021channel}. Many existing methods are supervised and rely on labeled data to achieve high performance. However, data labeling for large-scale datasets is extremely time-consuming and labor-intensive, motivating researchers to design self-supervised methods to reduce dependency on labeled data.

Reconstruction and contrastive learning are two primary pretext tasks in self-supervised skeleton-based human action recognition. Reconstruction methods learn features by predicting or reconstructing input data \cite{zheng2018unsupervised, kundu2019unsupervised, su2020predict}, whereas contrastive learning methods learn features by pulling positive sample pairs together and pushing negative sample pairs apart in the latent space \cite{rao2021augmented}. Recently, a few contrastive learning models such as SkeletonCLR \cite{li20213d}, AimCLR \cite{guo2022contrastive}, and ActCLR \cite{lin2023actionlet} have achieved high classification accuracy across various datasets. However, their adversarial robustness has not been thoroughly investigated.

In real-world scenarios where privacy or data transfer cost is a concern, edge devices usually extract skeleton data locally and only transfer the skeleton data to the cloud for human action recognition. In such distributed systems, the skeleton data is exposed and vulnerable during transfer. In this work, we show that the aforementioned mainstream self-supervised models lack robustness against adversarial attacks, where subtle, targeted perturbations to input data can significantly degrade their performance.

In general, defenses against adversarial attacks can be broadly categorized into data augmentation defenses \cite{madry2017towards, cohen2019certified}, model robustification defenses \cite{cisse2017parseval}, and hybrid defenses \cite{xie2019feature, zhang2019theoretically}. Data augmentation defenses enhance robustness by introducing perturbations during the training process (\eg, adversarial training and randomized smoothing), while model robustification defenses aim to design specific model architectures so that the model is naturally robust. Overall, data augmentation defenses achieve higher robustness, but rely on supervised learning or tailored contrastive learning frameworks \cite{kim2020adversarial, fan2021does}. In contrast, model robustification defenses have no such dependencies, but are generally more vulnerable to attacks. To the best of our knowledge, there is no specifically designed defense for self-supervised skeleton-based human action recognition.

Due to the research gap identified above, currently, the only potential solution is to adopt either general defenses such as TRADES \cite{zhang2019theoretically} or supervised defenses designed for skeleton-based human action recognition, such as BEAT \cite{wang2023defending}. However, applying these defenses to mainstream contrastive learning models, which represent the dominant pretext task in self-supervised skeleton-based human action recognition, yields limited robustness improvements. These observations highlight a critical need for a new defense tailored for these contrastive learning models. Ideally, such a solution should be a plug-in module that can be integrated into existing models without altering their primary training objectives.

To this end, we propose a plug-in robustification module via quantization for contrastive learning in self-supervised skeleton-based human action recognition (QCLR). Unlike most defenses that rely on data augmentation, QCLR falls into the less explored category of model robustification defenses, deriving robustness from its architectural design. To effectively defend against both gradient-based white-box and gradient-free black-box attacks, we introduce a dual-level quantization scheme across the encoding pipeline. By mapping continuous input and features into discrete spaces, this design creates a comprehensive defensive barrier: it absorbs adversarial perturbations within the quantization bins, while simultaneously breaking the gradient propagation. However, applying vector quantization inevitably introduces quantization error \cite{van2017neural}. As will be demonstrated in our experiments, applying naive vector quantization to mainstream contrastive learning models will severely degrade the representation quality, leading to a substantial decrease in the classification accuracy of the models.

To overcome this challenge, we propose a multi-faceted quantization strategy guided by an attention mechanism. Specifically, after conjecturing that the main source of quantization error is the limited codebook size, we introduce multi-faceted quantization (\ie, quantizing the continuous features of multiple facets, namely the spatial, temporal, and spatio-temporal facets). This quantization strategy essentially combines codewords from multiple codebooks to approximate the continuous feature vectors, thereby markedly enhancing the representational capability of discrete representations while maintaining computational tractability, despite the vast number of such combinations of codewords. Furthermore, we constrain the quantization processes by incorporating an attention mechanism. In the presence of multiple codebooks, each capturing one facet of the data, the quantization processes are underdetermined (\ie, different combinations of codewords can equally approximate the same continuous feature vector) and cause optimization difficulties. Therefore, the attention mechanism designates one facet as the anchor for the other facets. These quantization processes not only mitigate the impact of adversarial perturbations by mapping similar continuous feature vectors to the same codewords, but also break the gradient propagation, jointly enhancing robustness against adversarial attacks. While vector quantization inevitably introduces quantization error, with element-wise quantization retaining fine-grained details and the attention mechanism dynamically modulating the features of the anchor facet, we ensure the model preserves high performance. As a result, QCLR maintains highly competitive classification accuracy, while achieving a considerable improvement in adversarial robustness, demonstrating clear advantages over existing defenses.

We evaluate QCLR on several benchmark datasets across SkeletonCLR, AimCLR, and ActCLR. Compared to existing defenses, QCLR substantially improves model robustness against both gradient-based white-box and gradient-free black-box adversarial attacks, while maintaining classification accuracy comparable to that of the corresponding original models.

In summary, our contributions are as follows:
\begin{enumerate}
\item We propose QCLR, a plug-in robustification module employing a dual-level quantization scheme, and demonstrate its integration into a progressive lineage of mainstream contrastive learning models (\ie, SkeletonCLR, AimCLR, and ActCLR).
\item We design an attention-guided multi-faceted quantization strategy that not only breaks the gradient propagation and discretizes the feature spaces to enhance adversarial robustness, but also effectively mitigates the impact of the inevitable quantization error introduced by vector quantization.
\item We comprehensively evaluate QCLR on several benchmark datasets, demonstrating that it outperforms existing defenses in defending against both gradient-based white-box and gradient-free black-box attacks, while preserving highly competitive classification accuracy.
\end{enumerate}

\section{Related Work}

\subsection{Skeleton-Based Human Action Recognition}

Skeleton-based human action recognition refers to the task of recognizing human actions through the analysis of skeleton data, which typically consist of 3D coordinates of joints. Due to the stability of skeleton data against environmental factors such as background noise, scale changes, and lighting variations, as well as the advances in motion capture technologies that enhance fidelity, skeleton-based human action recognition has attracted significant interest in recent years.

Early methods rely on hand-crafted features to capture specific geometric or statistical patterns \cite{ellis2013exploring, muller2005efficient}. However, with the development of deep learning, learned features have become dominant. Deep learning network architectures learn representations without the need for manual feature engineering, excel at capturing complex, high-dimensional patterns, and are adaptable to diverse scenarios, leading to more accurate and reliable human action recognition.

Skeleton-based human action recognition methods using learned features can be broadly divided into three categories based on the deep learning network architectures they employ: convolutional neural network (CNN)-based, recurrent neural network (RNN)-based, and graph convolutional network (GCN)-based methods. CNN-based methods typically transform skeleton data into 2D matrices. Specifically, the 3D coordinates of joints in each frame are flattened into a vector. These vectors are then stacked together along the temporal dimension to form a spatial-temporal matrix, where one dimension represents frames and the other represents the flattened 3D coordinates of joints. This transformation enables convolution operations to simultaneously capture both spatial structures and short-term temporal patterns \cite{soo2017interpretable, du2015skeleton, ji20123d}. RNN-based methods focus on modeling the temporal dynamics of skeleton sequences by treating joint trajectories as time series and leveraging memory mechanisms to capture long-term dependencies across frames, enabling them to understand the temporal context of actions \cite{liu2016spatio, wang2017modeling, zheng2018unsupervised, kundu2019unsupervised, su2020predict}. GCN-based methods exploit the natural graph structure of skeleton data by representing joints as nodes and bones as edges. These methods model spatial dependencies between joints in each frame through graph convolution operations \cite{shi2019two, liu2020disentangling, chen2021channel}. Building upon the foundation of the GCN, the Spatial-Temporal Graph Convolutional Network (ST-GCN) further improves the capability of capturing temporal dependencies across frames by integrating temporal information into spatial graph structures, making it particularly effective in handling the dynamic patterns of skeleton data \cite{yan2018spatial}.

The design of our QCLR draws inspiration from these methods. Specifically, functioning as a plug-in module, QCLR incorporates a GCN to model spatial dependencies and a temporal CNN (T-CNN) to extract temporal patterns. These two parallel branches work in tandem with the standard ST-GCN, which is typically employed by mainstream self-supervised skeleton-based human action recognition models (\ie, SkeletonCLR, AimCLR, and ActCLR) to capture spatio-temporal relationships, generating the final integrated representations.

\subsection{Self-Supervised Skeleton-Based Human Action Recognition}

Self-supervised learning has emerged as a promising paradigm in skeleton-based human action recognition, enabling representation learning from unlabeled data. Unlike supervised approaches that achieve high performance but rely heavily on labeled data, self-supervised approaches address the challenges posed by time-consuming and labor-intensive data labeling. In this context, various self-supervised skeleton-based human action recognition methods have been developed.

In self-supervised skeleton-based human action recognition, pretext tasks play a crucial role in driving representation learning. Most existing methods adopt one of two primary pretext tasks: reconstruction and contrastive learning.

Reconstruction methods aim to learn informative representations through reconstructing or predicting the input skeleton data. Zheng et al. \cite{zheng2018unsupervised} integrated an autoencoder (AE) with a generative adversarial network (GAN), where the discriminator of the GAN enhances the performance of the decoder of the AE by enforcing realism in the reconstructed regions of the skeleton data. This method adopts the AE to drive representation learning, while using the GAN to ensure the spatial and temporal coherence within the reconstructed regions. Similarly, Kundu et al. \cite{kundu2019unsupervised} proposed a GAN with an AE setup specifically designed for constraining the representations, enabling the decoder of the AE to generate fake skeleton data, which the discriminator of the GAN evaluates against the real skeleton data to guide the AE accordingly. Su et al. \cite{su2020predict} improved the AE by introducing tailored decoder training strategies, which impose additional penalties on the decoder, guiding the encoder to learn more effective representations.

Contrastive learning methods focus on learning discriminative representations by contrasting positive and negative sample pairs. Most methods employ the contrastive learning framework of Momentum Contrast (MoCo) \cite{he2020momentum}, which leverages a dynamic queue to provide a large number of negative key features and a momentum update mechanism to ensure the stability of the key encoder, while incorporating their specific designs. Rao et al. \cite{rao2021augmented} proposed AS-CAL, which adopts the long short-term memory (LSTM) as the encoder. Li et al. \cite{li20213d} proposed SkeletonCLR, which uses the ST-GCN as the encoder, and CrosSCLR, which further extends SkeletonCLR with a cross-view knowledge mining strategy. Guo et al. \cite{guo2022contrastive} proposed AimCLR, which uses extreme data augmentation transformations and the energy-based attention-guided drop module (EADM) to produce diverse query features, along with Nearest Neighbors Mining (NNM) to further expand the set of positive key features, making the learning process more reliable. In addition, Lin et al. \cite{lin2023actionlet} proposed ActCLR, which decomposes action regions into actionlets and non-actionlets, adopting an adaptive transformation strategy to learn better semantic consistency from actionlets and obtain stronger generalization performance in non-actionlets.

Although recent works, including SkeletonCLR, AimCLR, and ActCLR, have achieved notable classification accuracy across various datasets, their vulnerability to adversarial attacks remains a significant concern. Our QCLR addresses this issue not as a standalone model, but as a plug-in module that can be integrated into these models, effectively mitigating the impact of adversarial perturbations while preserving highly competitive classification accuracy.

\section{Methodology}

Mainstream contrastive learning models for self-supervised skeleton-based human action recognition have proven highly effective in learning discriminative representations from unlabeled data, yet the continuous nature of these representations leaves them vulnerable to diverse adversarial attacks. To robustify existing models without altering their primary training objectives or relying on supervised learning, we employ a dual-level quantization scheme across the encoding pipeline, which introduces both input quantization and feature quantization, breaking the gradient propagation and discretizing the input and feature spaces to jointly achieve robustness against adversarial attacks.

To discretize the input and feature spaces, we employ vector quantization and element-wise quantization. However, naive vector quantization, which typically adopts a single codebook, exhibits severe representation degradation stemming from insufficient codebook size, leading to a significant decrease in classification accuracy. In the attempt to close the performance gap with continuous representations, the codebook size quickly grows beyond a feasible range, resulting in prohibitive memory cost and optimization difficulties. Therefore, simply scaling up the codebook size is not a practical solution.

To overcome this limitation, at the feature quantization level, we design a multi-faceted quantization strategy guided by an attention mechanism. This strategy adopts multiple codebooks, each corresponding to a distinct semantic facet (\ie, spatial, temporal, or spatio-temporal), significantly increasing the representational capability of discrete representations. Nevertheless, the presence of multiple codebooks introduces ambiguity since different combinations of codewords can equally approximate the same continuous feature vector, causing optimization difficulties. To address this ambiguity, we incorporate an attention mechanism that designates the temporal facet as the anchor and dynamically modulates its features with features of the spatial and spatio-temporal facets.

\begin{figure}[t]
\centering
\includegraphics[width=\textwidth, trim=0 113pt 0 113pt, clip]{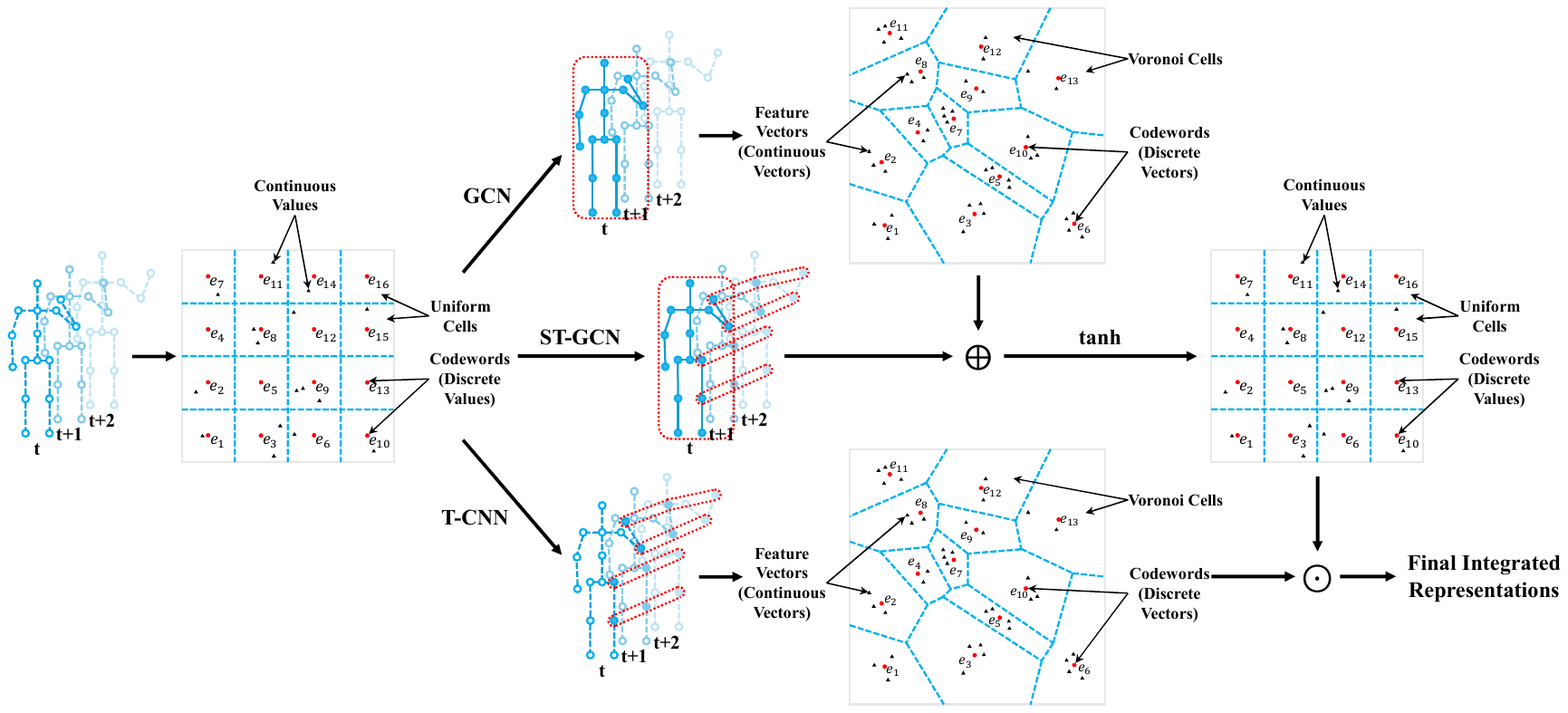}
\caption{An overview of QCLR. Note that the three parallel branches extract spatial (top), spatio-temporal (middle), and temporal (bottom) features, respectively.}
\label{fig:figure_1}
\vspace{-12pt}
\end{figure}

Thus, we propose QCLR, a plug-in robustification module that can be integrated into an evolving series of mainstream models (\ie, SkeletonCLR, AimCLR, and ActCLR) to enhance adversarial robustness while maintaining classification accuracy comparable to that of the corresponding original models. An overview of QCLR is illustrated in Figure~\ref{fig:figure_1}.

\subsection{Pipeline: A Dual-Level Quantization Scheme}

Functioning as a plug-in module, the proposed QCLR operates when integrated into mainstream contrastive learning models (\ie, SkeletonCLR, AimCLR, and ActCLR). To capture distinct semantic facets, QCLR introduces a GCN for spatial dependency modeling and a T-CNN for temporal pattern extraction. These parallel branches work in tandem with the standard ST-GCN employed by these models. Overall, QCLR strategically operates around the ST-GCN and strictly retains the data augmentation transformations, the projection head, and the contrastive objective.

The input of QCLR is a skeleton sequence, represented as $\mathcal{X} \in \mathbb{R}^{3 \times T \times V}$, where $3$, $T$, and $V$ denote the 3D coordinates of joints, the number of frames (\ie, the temporal dimension), and the number of joints (\ie, the spatial dimension), respectively. In the first level of our dual-level quantization scheme, the continuous input $\mathcal{X}$ undergoes element-wise quantization using a fixed codebook $\mathcal{C}^x = \{e_i^x\}_{i=1}^{N^x}$, $e_i^x \in \mathbb{R}$ to yield the quantized input:

\begin{equation}
\mathcal{X}_q = \operatorname{Q}(\mathcal{X}, \mathcal{C}^x)
\end{equation}

The resulting quantized input $\mathcal{X}_q$ is processed by three parallel branches to extract features from distinct semantic facets: (1) the GCN models intra-frame spatial dependencies to extract spatial features $\mathcal{Z}^s \in \mathbb{R}^{D \times T \times V}$, where $D$ represents the feature dimension; (2) the ST-GCN captures spatio-temporal relationships to produce spatio-temporal features $\mathcal{Z}^{st} \in \mathbb{R}^{D \times T' \times V}$, where $T'$ denotes the temporal dimension after downsampling; (3) the T-CNN extracts inter-frame temporal patterns to yield temporal features $\mathcal{Z}^t \in \mathbb{R}^{D \times T' \times V}$. Subsequently, both $\mathcal{Z}^s$ and $\mathcal{Z}^t$ undergo vector quantization using trainable codebooks $\mathcal{C}^s = \{e_i^s\}_{i=1}^{N^s}$, $e_i^s \in \mathbb{R}^D$ and $\mathcal{C}^t = \{e_i^t\}_{i=1}^{N^t}$, $e_i^t \in \mathbb{R}^D$, respectively, to obtain the quantized spatial and temporal features:

\begin{equation}
\mathcal{Z}_q^s = \operatorname{VQ}(\mathcal{Z}^s, \mathcal{C}^s), \quad
\mathcal{Z}_q^t = \operatorname{VQ}(\mathcal{Z}^t, \mathcal{C}^t)
\end{equation}

Next, the quantized spatial features $\mathcal{Z}_q^s$ are temporally aligned with and then combined with the spatio-temporal features $\mathcal{Z}^{st}$, followed by the $\operatorname{tanh}$ activation function to produce the attention weights $\mathcal{W}$. The weights then undergo element-wise quantization using a fixed codebook $\mathcal{C}^w = \{e_i^w\}_{i=1}^{N^w}$, $e_i^w \in \mathbb{R}$ to derive the quantized attention weights:

\begin{equation}
\mathcal{W}_q = \operatorname{Q}(\mathcal{W}, \mathcal{C}^w)
\end{equation}

The quantized attention weights $\mathcal{W}_q$ dynamically modulate the quantized temporal features $\mathcal{Z}_q^t$, resulting in the final integrated representations $\mathcal{Z}_{f}$.

The final integrated representations $\mathcal{Z}_{f}$ undergo spatio-temporal average pooling, eliminating the spatial and temporal dimensions, prior to being averaged along the subject dimension and then passed through the retained projection head for contrastive learning. The loss function is composed of the contrastive loss $\mathcal{L}_c$ derived from the original contrastive learning model, which pulls positive sample pairs together and pushes negative sample pairs apart in the latent space, and the vector quantization losses \cite{van2017neural} to mutually align the continuous feature vectors and their corresponding codewords. Specifically, the loss function is formulated as:

\begin{equation}
\mathcal{L} = \mathcal{L}_c + \mathcal{L}^s_q + \mathcal{L}^t_q
\end{equation}

where $\mathcal{L}^s_q$ and $\mathcal{L}^t_q$ are the vector quantization losses for the spatial and temporal features, respectively.

\subsection{Input Quantization}

Operating on the continuous input $\mathcal{X}$ immediately upon it entering the encoder, input quantization serves as the first level of the proposed dual-level quantization scheme. Conceptually, the goal is to map each continuous input value to its nearest codeword within a uniformly spaced codebook $\mathcal{C}^x$ consisting of discrete values with a step size of $0.1$. Since the predefined codebook is uniformly spaced and fixed, the element-wise quantization operation can be formulated as a closed-form expression. Specifically, without requiring explicit nearest codeword search, the quantized input $\mathcal{X}_q$ is derived via a rounding operation:

\begin{equation}
\mathcal{X}_q = \operatorname{round}(\mathcal{X} \times 10) \times 10^{-1}
\end{equation}

By discretizing the input space and breaking the gradient propagation, this quantization process effectively mitigates the impact of small adversarial perturbations. Due to its non-differentiable nature, a straight-through estimator (STE) \cite{bengio2013estimating} is employed to estimate gradients during backpropagation. Notably, input quantization introduces no additional loss and is applied consistently across all stages, namely pretext training, linear evaluation, and inference.

\subsection{Attention-Guided Multi-Faceted Feature Quantization}

Serving as the second level of the proposed dual-level quantization scheme, feature quantization applies a multi-faceted quantization strategy guided by an attention mechanism to the features extracted by the three parallel branches, resulting in the final integrated representations. In particular, this tailored strategy uses a combination of vector quantization and element-wise quantization.

For the spatial and temporal branches (\ie, the GCN and T-CNN), vector quantization performs explicit nearest codeword search for each continuous feature vector. Specifically, given a feature vector $z$, it is mapped to its nearest codeword within the corresponding trainable codebook:

\begin{equation}
z_q = e_k \text{, where } k = \operatorname*{argmin}_j \lVert z - e_j \rVert
\end{equation}

where $z_q$ is the quantized feature vector, $e_j$ is the $j$-th codeword within the codebook, and $k$ is the index of the nearest codeword to $z$.

To address the ambiguity introduced by the presence of multiple codebooks, we employ an attention mechanism to integrate the features from all three parallel branches. Specifically, the quantized spatial features $\mathcal{Z}_q^s$ are first downsampled by a linear layer, denoted as $\phi(\cdot)$, to be temporally aligned with the spatio-temporal features $\mathcal{Z}^{st}$. Then, the most informative spatio-temporal features are added to these severely degraded quantized spatial features, thereby compensating for the representation degradation, prior to being passed through the $\operatorname{tanh}$ activation function to yield the attention weights:

\begin{equation}
\mathcal{W} = \operatorname{tanh}(\phi(\mathcal{Z}_q^s) + \mathcal{Z}^{st})
\end{equation}

Subsequently, the attention weights are subjected to element-wise quantization using a fixed, uniformly spaced codebook $\mathcal{C}^w$ with a step size of $0.001$. Similar to input quantization, this operation can also be formulated as a closed-form expression, in which the quantized attention weights $\mathcal{W}_q$ are derived via a rounding operation:

\begin{equation}
\mathcal{W}_q = \operatorname{round}(\mathcal{W} \times 10^3) \times 10^{-3}
\end{equation}

Finally, these quantized attention weights dynamically modulate the quantized temporal features $\mathcal{Z}_q^t$, which are entirely composed of temporal codewords and are thus the most suitable base features, through element-wise multiplication to obtain the final integrated representations:

\begin{equation}
\mathcal{Z}_f = \mathcal{W}_q \odot \mathcal{Z}_q^t
\end{equation}

By discretizing the feature spaces and breaking the gradient propagation, these quantization processes significantly mitigate the impact of adversarial perturbations. Although vector quantization inevitably introduces quantization error, element-wise quantization and the attention mechanism compensate for it, thereby preserving highly competitive classification accuracy. Furthermore, due to the non-differentiable nature of the quantization operations, STEs are employed to estimate gradients during backpropagation.

\section{Experiments}

\subsection{Datasets}

For evaluation, we conduct our experiments on the following three datasets:

The NTU RGB+D 60 dataset (NTU-60) \cite{shahroudy2016ntu} is a large-scale benchmark dataset widely used for skeleton-based human action recognition, consisting of 56,578 samples across 60 action classes. Each sample is a temporal sequence of skeleton graphs, with each graph comprising 3D coordinates of 25 joints.

The NTU RGB+D 120 dataset (NTU-120) \cite{liu2019ntu} is the extended version of NTU-60. As one of the largest benchmark datasets for human action recognition, it comprises 113,945 samples across 120 action classes.

The PKU Multi-Modality dataset (PKU-MMD) \cite{liu2020benchmark} is a large-scale benchmark dataset for multi-modality human action analysis. It consists of two parts, Part I and Part II. In our experiments, we use Part I, which contains 21,539 samples across 51 action classes.

\subsection{Experimental Settings}

To comprehensively evaluate the proposed plug-in robustification module, we integrate QCLR into SkeletonCLR, AimCLR, and ActCLR, resulting in three model variants: Q-SkeletonCLR, Q-AimCLR, and Q-ActCLR.

The GCN consists of 6 GCN units, which are derived directly from the standard ST-GCN unit by retaining the graph convolution component while entirely removing the temporal convolution component. The spatial kernel size of each GCN unit is kept identical to that of the standard ST-GCN unit and the numbers of output channels of the 6 GCN units are configured as $64$, $64$, $128$, $128$, $256$, and $256$, respectively. The strides of all 6 GCN units are set to $1$.

The T-CNN consists of 4 T-CNN units, which are constructed by removing the graph convolution component from the standard ST-GCN unit, as well as the first batch normalization, the ReLU, and the dropout layers of the temporal convolution component. The temporal kernel size of each T-CNN unit remains identical to that of the standard ST-GCN unit and the numbers of output channels of the 4 T-CNN units are configured as $64$, $128$, $256$, and $256$, respectively. The strides of the second and third T-CNN units are set to $2$ for temporal downsampling, whereas those of the first and fourth T-CNN units are set to $1$.

Consistent with the standard ST-GCN, the first units of both the GCN and T-CNN exclude the residual connections. For vector quantization, the spatial and temporal codebooks contain 512 and 256 codewords, respectively.

We follow the data preprocessing pipeline shared by all three original models to ensure a fair comparison. We conduct all experiments under the cross-subject setting, where the training and test data are collected from different subjects.

During both the pretext training and linear evaluation stages, the training and evaluation settings of all model variants, including hyperparameters such as the number of epochs, the optimizer configurations, and the learning rate schedules, strictly follow those of their corresponding original models. In the linear evaluation stage, all parameters within the query encoder of each model variant are frozen. The final integrated representations are flattened across the feature, temporal, and spatial dimensions, and then averaged along the subject dimension, before being passed through a linear classifier, which is a fully connected layer (\ie, linear layer) followed by a softmax layer.

For each model, the number of parameters increases from 1.84M to 5.12M, and the time for pretext training and linear evaluation triples. Although the time for inference also triples, a speed of 433 samples per second remains practically efficient. Regarding memory cost, a single TITAN RTX is sufficient.

\subsection{Attack and Defense Settings}

Following the linear evaluation stage, we collect correctly classified test samples to evaluate adversarial robustness. Given that the quantization operations are non-differentiable, gradient-based white-box attacks cannot be directly performed. Therefore, we implement Backward Pass Differentiable Approximation (BPDA) \cite{athalye2018obfuscated} via STEs to estimate gradients during backpropagation.

We employ three gradient-based white-box attacks: PGD \cite{madry2017towards}, a general and widely used benchmark attack for assessing model robustness; SMART \cite{wang2021understanding}, an attack specifically designed for skeleton-based human action recognition; and APGD-DLR \cite{croce2020reliable}, an attack that adopts the difference of logits ratio loss to obtain more stable and informative gradients, thereby bypassing gradient obfuscation defenses. PGD and SMART are performed for 300 iterations, with PGD configured with a perturbation budget of $0.01$ and a step size of $0.04/300$. APGD-DLR is performed for 100 iterations with a perturbation budget of $0.005$.

Additionally, we use BASAR \cite{diao2021basar}, a gradient-free black-box attack specifically designed for skeleton-based human action recognition. BASAR is performed for 500 iterations on 1,000 randomly selected, correctly classified test samples. As BASAR always achieves a 100\% attack success rate (ASR) by design, we assess the quality of the generated adversarial samples using three metrics: the averaged joint position deviation $l$, the averaged joint acceleration deviation $\Delta a$, and the averaged bone-length deviation percentage $\Delta B/B$. These metrics collectively reflect the imperceptibility of the attack, where larger values generally indicate noticeable skeleton jitter or unnatural skeleton deformation.

We adopt two baseline defenses: TRADES \cite{zhang2019theoretically}, a general adversarial defense; and BEAT \cite{wang2023defending}, a supervised defense tailored for skeleton-based human action recognition. TRADES is performed for 10 iterations, with a trade-off regularization hyperparameter of $1$, a perturbation budget of $0.01$, and a step size of $0.04/10$. BEAT is configured with 5 appended models.

\begin{table}[t]
\centering
\resizebox{\textwidth}{!}{
\begin{tabular}{c cccc ccc cccc ccc cccc ccc}
\toprule
\multirow{3}{*}[-1.05ex]{Model} & \multicolumn{7}{c}{NTU-60} & \multicolumn{7}{c}{NTU-120} & \multicolumn{7}{c}{PKU-MMD} \\
\cmidrule(lr){2-8} \cmidrule(lr){9-15} \cmidrule(lr){16-22}
 &
 \multirow{2}{*}[-0.525ex]{\makecell[c]{Classification \\ Accuracy}} & \multirow{2}{*}[-0.525ex]{\makecell[c]{PGD \\ ASR}} & \multirow{2}{*}[-0.525ex]{\makecell[c]{SMART \\ ASR}} & \multirow{2}{*}[-0.525ex]{\makecell[c]{APGD-DLR \\ ASR}} & \multicolumn{3}{c}{BASAR} &
 \multirow{2}{*}[-0.525ex]{\makecell[c]{Classification \\ Accuracy}} & \multirow{2}{*}[-0.525ex]{\makecell[c]{PGD \\ ASR}} & \multirow{2}{*}[-0.525ex]{\makecell[c]{SMART \\ ASR}} & \multirow{2}{*}[-0.525ex]{\makecell[c]{APGD-DLR \\ ASR}} & \multicolumn{3}{c}{BASAR} &
 \multirow{2}{*}[-0.525ex]{\makecell[c]{Classification \\ Accuracy}} & \multirow{2}{*}[-0.525ex]{\makecell[c]{PGD \\ ASR}} & \multirow{2}{*}[-0.525ex]{\makecell[c]{SMART \\ ASR}} & \multirow{2}{*}[-0.525ex]{\makecell[c]{APGD-DLR \\ ASR}} & \multicolumn{3}{c}{BASAR} \\
\cmidrule(lr){6-8} \cmidrule(lr){13-15} \cmidrule(lr){20-22}
 &
 & & & & $l$ & $\Delta a$ & $\Delta B/B$ &
 & & & & $l$ & $\Delta a$ & $\Delta B/B$ &
 & & & & $l$ & $\Delta a$ & $\Delta B/B$ \\
\midrule
SkeletonCLR &
68.3 & 99.98 & 100 & 99.42 & 0.09 & 0.03 & 8.5 &
56.8 & 100 & 100 & 99.93 & 0.08 & 0.02 & 5.0 &
80.9 & 99.82 & 100 & 98.67 & 0.09 & 0.03 & 8.2 \\
TRADES-SkeletonCLR &
63.83 & 93.96 & 99.99 & 83.94 & 0.15 & \textbf{0.05} & \textbf{17.4} &
49.66 & 99.67 & 99.91 & 98.48 & 0.12 & 0.03 & 10.9 &
76.07 & 93.73 & 100 & 80.85 & 0.17 & \textbf{0.05} & 16.6 \\
BEAT-SkeletonCLR &
69.56 & 97.96 & 94.03 & 99.37 & 0.09 & 0.03 & 9.1 &
57.04 & 99.30 & 97.66 & 99.94 & 0.07 & 0.02 & 4.9 &
79.84 & 93.84 & 85.83 & 98.75 & 0.09 & 0.03 & 8.0 \\
\textbf{Q-SkeletonCLR (Ours)} &
\textbf{71.02} & \textbf{5.99} & \textbf{4.13} & \textbf{54.86} & \textbf{0.36} & 0.04 & 15.4 &
\textbf{60.63} & \textbf{5.19} & \textbf{3.24} & \textbf{63.74} & \textbf{0.32} & \textbf{0.04} & \textbf{14.5} &
\textbf{83.21} & \textbf{3.29} & \textbf{2.04} & \textbf{30.93} & \textbf{0.37} & \textbf{0.05} & \textbf{16.9} \\
\midrule
AimCLR &
74.3 & 99.97 & 100 & 97.64 & 0.10 & 0.04 & 11.7 &
\textbf{63.4} & 99.99 & 100 & 98.99 & 0.09 & 0.03 & 10.3 &
83.4 & 100 & 100 & 94.48 & 0.11 & 0.04 & 13.4 \\
TRADES-AimCLR &
69.38 & 93.51 & 100 & 82.45 & 0.16 & \textbf{0.06} & \textbf{20.2} &
54.35 & 97.86 & 100 & 88.55 & 0.14 & \textbf{0.05} & \textbf{16.4} &
79.62 & 91.41 & 100 & 71.71 & 0.19 & \textbf{0.07} & \textbf{22.8} \\
BEAT-AimCLR &
\textbf{74.76} & 94.69 & 77.31 & 97.38 & 0.10 & 0.04 & 11.6 &
62.31 & 96.98 & 84.87 & 98.83 & 0.09 & 0.03 & 9.7 &
\textbf{84.13} & 95.65 & 77.45 & 94.29 & 0.11 & 0.04 & 13.0 \\
\textbf{Q-AimCLR (Ours)} &
73.77 & \textbf{8.07} & \textbf{4.79} & \textbf{58.93} & \textbf{0.41} & \textbf{0.06} & 18.6 &
61.66 & \textbf{12.92} & \textbf{8.39} & \textbf{66.54} & \textbf{0.34} & 0.04 & 14.1 &
82.99 & \textbf{3.74} & \textbf{1.65} & \textbf{31.60} & \textbf{0.37} & 0.05 & 16.5 \\
\midrule
ActCLR &
\textbf{80.9} & 99.96 & 99.71 & 100 & 0.12 & 0.03 & 10.9 &
\textbf{69.0} & 99.90 & 99.84 & 99.93 & 0.12 & 0.03 & 9.1 &
89.68 & 99.63 & 100 & 88.54 & 0.16 & 0.05 & 14.9 \\
TRADES-ActCLR &
78.12 & 100 & 99.96 & 99.96 & 0.17 & 0.05 & 16.2 &
63.76 & 99.10 & 99.82 & 97.82 & 0.16 & 0.04 & 14.0 &
87.06 & 92.52 & 100 & 67.25 & 0.22 & \textbf{0.07} & \textbf{22.1} \\
BEAT-ActCLR &
80.31 & 93.95 & 73.22 & 100 & 0.12 & 0.03 & 10.0 &
68.99 & 94.60 & 79.95 & 99.87 & 0.12 & 0.03 & 8.9 &
\textbf{89.72} & 86.48 & 71.27 & 88.71 & 0.15 & 0.05 & 15.0 \\
\textbf{Q-ActCLR (Ours)} &
77.47 & \textbf{1.30} & \textbf{0.84} & \textbf{41.40} & \textbf{0.43} & \textbf{0.06} & \textbf{20.4} &
65.87 & \textbf{1.79} & \textbf{1.18} & \textbf{41.89} & \textbf{0.40} & \textbf{0.05} & \textbf{17.9} &
87.68 & \textbf{2.49} & \textbf{1.48} & \textbf{37.62} & \textbf{0.39} & 0.05 & 17.9 \\
\bottomrule
\end{tabular}
}
\caption{Results of adversarial robustness evaluation across three datasets. Evaluation metrics include the classification accuracy (\%, $\uparrow$), the ASR (\%, $\downarrow$) of PGD, SMART, and APGD-DLR, and the three metrics ($l$ ($\uparrow$), $\Delta a$ ($\uparrow$), $\Delta B/B$ (\%, $\uparrow$)) of BASAR. The best results for each model are indicated in \textbf{bold}.}
\label{tab:table_1}
\vspace{-23pt}
\end{table}

\subsection{Evaluation and Comparison}

The comprehensive evaluation results of adversarial robustness across the three datasets are summarized in Table~\ref{tab:table_1}. We compare the proposed QCLR with baseline defenses on three mainstream contrastive learning models.

A fundamental challenge in adversarial defense lies in the trade-off between classification accuracy and model robustness. Experimental results show that adopting the general defense TRADES introduces a noticeable decrease in classification accuracy. In contrast, BEAT, which is tailored for skeleton-based human action recognition, effectively preserves model performance. As for QCLR, it maintains highly competitive classification accuracy, despite exhibiting slightly larger variations compared to BEAT. Notably, Q-SkeletonCLR achieves higher classification accuracy than SkeletonCLR across all three datasets.

When subjected to gradient-based white-box attacks, the baseline defenses demonstrate remarkably limited defensive capabilities. Empirical results indicate that TRADES offers only partial protection against PGD and APGD-DLR, yet fails to withstand SMART. Similarly, while BEAT shows some effectiveness against SMART, it remains highly vulnerable to PGD and APGD-DLR. By contrast, QCLR not only substantially lowers the ASR of PGD and SMART to single digits in almost all scenarios, but also consistently maintains the lowest ASR of APGD-DLR, which is designed to bypass gradient obfuscation defenses and provide more reliable robustness evaluations. This suggests that QCLR derives its robustness from the architectural design, absorbing small perturbations within the quantization bins rather than merely obfuscating gradients.

Against the gradient-free black-box attack BASAR, which naturally bypasses gradient obfuscation defenses, QCLR demonstrates significant robustness by forcing a substantial degradation in the quality of generated adversarial samples. While TRADES also exhibits notable defensive capability, QCLR forces markedly high values across all three metrics of BASAR, particularly for the averaged joint position deviation $l$. The increase in these metrics indicates more severe jitter and deformation within the adversarial samples, making them likely to be perceptible. This further confirms that QCLR achieves robustness by absorbing adversarial perturbations within the quantization bins, rather than through mere gradient obfuscation.

\subsection{Ablation Study}

To verify the effectiveness of the components of the proposed QCLR, we conduct ablation experiments on NTU-60, the most widely used large-scale benchmark dataset for skeleton-based human action recognition. First, we demonstrate the necessity of the attention-guided multi-faceted quantization strategy by revealing the representational limitations of naive vector quantization. Second, we decompose the dual-level quantization scheme to analyze the complementary defensive capabilities of input quantization and feature quantization. Finally, we adjust the step size of input quantization to explore the trade-off between classification accuracy and adversarial robustness.

\vspace{-10pt}
\begin{table}[h]
\centering
\resizebox{\textwidth}{!}{
\begin{tabular}{c c c c c}
\toprule
Model & \makecell[c]{Classification \\ Accuracy} & \makecell[c]{Classification Accuracy with \\ the First Implementation} & \makecell[c]{Classification Accuracy with \\ the Second Implementation} & \makecell[c]{Classification Accuracy with \\ Feature Quantization (Ours)} \\
\midrule
SkeletonCLR & 68.3 & 1.67 & 42.55 & \textbf{72.95} \\
AimCLR & 74.3 & 1.67 & 48.65 & \textbf{76.07} \\
ActCLR & \textbf{80.9} & 1.67 & 64.55 & 80.85 \\
\bottomrule
\end{tabular}
}
\caption{Results of different vector quantization strategies on NTU-60. The best result for each model is indicated in \textbf{bold}.}
\label{tab:table_2}
\vspace{-26pt}
\end{table}

We compare our tailored feature quantization strategy with two implementations of naive vector quantization, each of which adopts a single codebook limited to 4096 codewords due to memory constraints. As shown in Table~\ref{tab:table_2}, the first implementation uses quantized features for both pretext training and linear evaluation, which causes the optimization process to diverge, resulting in the complete failure of contrastive learning and near-random classification accuracy. The second implementation uses continuous features during pretext training while simultaneously training the codebook, reserving quantized features only for linear evaluation. Although this implementation stabilizes the contrastive learning process, it still leads to a significant degradation in classification accuracy. Theoretically, only when the codebook becomes sufficiently large that every distinct continuous feature vector has a distinct corresponding codeword can the classification accuracy reach that of the original models. By contrast, our attention-guided multi-faceted feature quantization strategy maintains highly competitive classification accuracy, confirming that it effectively mitigates the impact of inevitable quantization error introduced by vector quantization.

\vspace{-10pt}
\begin{table}[h]
\centering
\resizebox{\textwidth}{!}{
\begin{tabular}{c cccc ccc cccc ccc cccc ccc}
\toprule
\multirow{3}{*}[-1.05ex]{Defense} & \multicolumn{7}{c}{SkeletonCLR} & \multicolumn{7}{c}{AimCLR} & \multicolumn{7}{c}{ActCLR} \\
\cmidrule(lr){2-8} \cmidrule(lr){9-15} \cmidrule(lr){16-22}
 &
 \multirow{2}{*}[-0.525ex]{\makecell[c]{Classification \\ Accuracy}} & \multirow{2}{*}[-0.525ex]{\makecell[c]{PGD \\ ASR}} & \multirow{2}{*}[-0.525ex]{\makecell[c]{SMART \\ ASR}} & \multirow{2}{*}[-0.525ex]{\makecell[c]{APGD-DLR \\ ASR}} & \multicolumn{3}{c}{BASAR} &
 \multirow{2}{*}[-0.525ex]{\makecell[c]{Classification \\ Accuracy}} & \multirow{2}{*}[-0.525ex]{\makecell[c]{PGD \\ ASR}} & \multirow{2}{*}[-0.525ex]{\makecell[c]{SMART \\ ASR}} & \multirow{2}{*}[-0.525ex]{\makecell[c]{APGD-DLR \\ ASR}} & \multicolumn{3}{c}{BASAR} &
 \multirow{2}{*}[-0.525ex]{\makecell[c]{Classification \\ Accuracy}} & \multirow{2}{*}[-0.525ex]{\makecell[c]{PGD \\ ASR}} & \multirow{2}{*}[-0.525ex]{\makecell[c]{SMART \\ ASR}} & \multirow{2}{*}[-0.525ex]{\makecell[c]{APGD-DLR \\ ASR}} & \multicolumn{3}{c}{BASAR} \\
\cmidrule(lr){6-8} \cmidrule(lr){13-15} \cmidrule(lr){20-22}
 &
 & & & & $l$ & $\Delta a$ & $\Delta B/B$ &
 & & & & $l$ & $\Delta a$ & $\Delta B/B$ &
 & & & & $l$ & $\Delta a$ & $\Delta B/B$ \\
\midrule
N/A &
68.3 & 99.98 & 100 & 99.42 & 0.09 & 0.03 & 8.5 &
74.3 & 99.97 & 100 & 97.64 & 0.10 & 0.04 & 11.7 &
\textbf{80.9} & 99.96 & 99.71 & 100 & 0.12 & 0.03 & 10.9 \\
Input Quantization &
70.64 & 84.46 & 100 & 59.93 & 0.35 & \textbf{0.05} & \textbf{17.1} &
73.24 & 81.66 & 97.42 & \textbf{54.46} & \textbf{0.43} & \textbf{0.06} & \textbf{20.6} &
76.27 & 72.71 & 99.57 & \textbf{41.27} & 0.41 & \textbf{0.06} & 19.7 \\
Feature Quantization &
\textbf{72.95} & \textbf{5.94} & \textbf{3.28} & 100 & 0.06 & 0.01 & 4.5 &
\textbf{76.07} & 9.74 & 5.31 & 97.49 & 0.17 & 0.04 & 12.7 &
80.85 & \textbf{1.06} & 1.09 & 100 & 0.11 & 0.03 & 8.9 \\
\textbf{QCLR} &
71.02 & 5.99 & 4.13 & \textbf{54.86} & \textbf{0.36} & 0.04 & 15.4 &
73.77 & \textbf{8.07} & \textbf{4.79} & 58.93 & 0.41 & \textbf{0.06} & 18.6 &
77.47 & 1.30 & \textbf{0.84} & 41.40 & \textbf{0.43} & \textbf{0.06} & \textbf{20.4} \\
\bottomrule
\end{tabular}
}
\caption{Results of different quantization levels of QCLR on NTU-60. The best results for each model are indicated in \textbf{bold}.}
\label{tab:table_3}
\vspace{-26pt}
\end{table}

To illustrate the complementary defensive capabilities of input quantization and feature quantization, we decompose the dual-level quantization scheme. As presented in Table~\ref{tab:table_3}, input quantization significantly lowers the ASR of APGD-DLR and forces the values across all three metrics of BASAR to increase, yet remains highly vulnerable to PGD and SMART. In contrast, feature quantization yields substantial robustness against PGD and SMART, but fails to withstand APGD-DLR and BASAR. Their complementary defensive capabilities can be attributed to their distinct operational levels. Input quantization discretizes the input space, hindering attacks that rely on subtle variations in perturbations for step-size adapting and boundary probing. Meanwhile, feature quantization discretizes the spatial, temporal, and spatio-temporal feature spaces, hindering conventional iterative attacks that depend on successive updates along the directions of the estimated gradients. By integrating both quantization levels, QCLR leverages their complementary advantages, achieving comprehensive robustness against diverse adversarial attacks.

\vspace{-10pt}
\begin{table}[h]
\centering
\resizebox{\textwidth}{!}{
\begin{tabular}{c cc ccc cc ccc cc ccc}
\toprule
\multirow{3}{*}[-1.05ex]{Step Size} & \multicolumn{5}{c}{SkeletonCLR} & \multicolumn{5}{c}{AimCLR} & \multicolumn{5}{c}{ActCLR} \\
\cmidrule(lr){2-6} \cmidrule(lr){7-11} \cmidrule(lr){12-16}
 &
 \multirow{2}{*}[-0.525ex]{\makecell[c]{Classification \\ Accuracy}} & \multirow{2}{*}[-0.525ex]{\makecell[c]{APGD-DLR \\ ASR}} & \multicolumn{3}{c}{BASAR} &
 \multirow{2}{*}[-0.525ex]{\makecell[c]{Classification \\ Accuracy}} & \multirow{2}{*}[-0.525ex]{\makecell[c]{APGD-DLR \\ ASR}} & \multicolumn{3}{c}{BASAR} &
 \multirow{2}{*}[-0.525ex]{\makecell[c]{Classification \\ Accuracy}} & \multirow{2}{*}[-0.525ex]{\makecell[c]{APGD-DLR \\ ASR}} & \multicolumn{3}{c}{BASAR} \\
\cmidrule(lr){4-6} \cmidrule(lr){9-11} \cmidrule(lr){14-16}
 &
 & & $l$ & $\Delta a$ & $\Delta B/B$ &
 & & $l$ & $\Delta a$ & $\Delta B/B$ &
 & & $l$ & $\Delta a$ & $\Delta B/B$ \\
\midrule
N/A &
68.3 & 99.42 & 0.09 & 0.03 & 8.5 &
74.3 & 97.64 & 0.10 & 0.04 & 11.7 &
\textbf{80.9} & 100 & 0.12 & 0.03 & 10.9 \\
0.01 &
\textbf{71.86} & 89.09 & 0.32 & \textbf{0.05} & 16.0 &
\textbf{75.85} & 93.52 & 0.28 & 0.05 & 17.7 &
78.72 & 93.29 & 0.29 & 0.05 & 16.9 \\
0.05 &
70.84 & 67.62 & \textbf{0.36} & \textbf{0.05} & \textbf{17.0} &
74.57 & 75.43 & 0.39 & \textbf{0.06} & \textbf{18.6} &
77.87 & 63.88 & 0.41 & \textbf{0.06} & 19.3 \\
\textbf{0.1} &
71.02 & \textbf{54.86} & \textbf{0.36} & 0.04 & 15.4 &
73.77 & \textbf{58.93} & \textbf{0.41} & \textbf{0.06} & \textbf{18.6} &
77.47 & \textbf{41.40} & \textbf{0.43} & \textbf{0.06} & \textbf{20.4} \\
\bottomrule
\end{tabular}
}
\caption{Results of different step sizes of input quantization on NTU-60. The best results for each model are indicated in \textbf{bold}.}
\label{tab:table_4}
\vspace{-26pt}
\end{table}

Furthermore, we demonstrate that the trade-off between classification accuracy and adversarial robustness against APGD-DLR and BASAR can be controlled by adjusting the step size of input quantization. As illustrated in Table~\ref{tab:table_4}, when the step size is reduced, the classification accuracy of model variants steadily improves, approaching or even surpassing that of the corresponding original models. Meanwhile, the defensive capability against both APGD-DLR and BASAR degrades: the ASR of APGD-DLR increases substantially, and all three metrics of BASAR decrease markedly. This consistent monotonic trend further confirms that the step size of input quantization determines the degree of discreteness of the input space, thereby influencing both classification accuracy and adversarial robustness. Thus, the step size of input quantization can serve as a hyperparameter to control the trade-off between classification accuracy and adversarial robustness.

\section{Conclusions}

In this work, we propose QCLR, a plug-in robustification module designed to address the vulnerability of mainstream contrastive learning models for self-supervised skeleton-based human action recognition against adversarial attacks. QCLR employs a dual-level quantization scheme that discretizes the input and feature spaces, breaking the gradient propagation and absorbing adversarial perturbations within the quantization bins. To overcome the severe representation degradation typically caused by vector quantization, we introduce an attention-guided multi-faceted quantization strategy, which effectively mitigates the impact of quantization error and preserves the representation quality. Extensive evaluations on three large-scale benchmark datasets demonstrate that integrating QCLR into a progressive lineage of mainstream models (\ie, SkeletonCLR, AimCLR, and ActCLR) yields comprehensive robustness against both gradient-based white-box and gradient-free black-box attacks. Moreover, although QCLR is developed for the ST-GCN-based models, its high-level design, which is to avoid applying vector quantization to the highly informative spatio-temporal features, does not depend on any specific network architecture. Thus, extending it to other network architectures remains our future direction.
\bibliographystyle{splncs04}
\bibliography{main}

@String(AAAI  = {AAAI})

@article{wang2013dense,
  title={Dense trajectories and motion boundary descriptors for action recognition},
  author={Wang, Heng and Kl{\"a}ser, Alexander and Schmid, Cordelia and Liu, Cheng-Lin},
  journal={International journal of computer vision},
  volume={103},
  pages={60--79},
  year={2013},
  publisher={Springer}
}

@inproceedings{luo2013group,
  title={Group sparsity and geometry constrained dictionary learning for action recognition from depth maps},
  author={Luo, Jiajia and Wang, Wei and Qi, Hairong},
  booktitle={Proceedings of the IEEE international conference on computer vision},
  pages={1809--1816},
  year={2013}
}

@inproceedings{hu2015jointly,
  title={Jointly learning heterogeneous features for RGB-D activity recognition},
  author={Hu, Jian-Fang and Zheng, Wei-Shi and Lai, Jianhuang and Zhang, Jianguo},
  booktitle={Proceedings of the IEEE conference on computer vision and pattern recognition},
  pages={5344--5352},
  year={2015}
}

@inproceedings{liu2016spatio,
  title={Spatio-temporal lstm with trust gates for 3d human action recognition},
  author={Liu, Jun and Shahroudy, Amir and Xu, Dong and Wang, Gang},
  booktitle={Computer Vision--ECCV 2016: 14th European Conference, Amsterdam, The Netherlands, October 11-14, 2016, Proceedings, Part III 14},
  pages={816--833},
  year={2016},
  organization={Springer}
}

@inproceedings{wang2017modeling,
  title={Modeling temporal dynamics and spatial configurations of actions using two-stream recurrent neural networks},
  author={Wang, Hongsong and Wang, Liang},
  booktitle={Proceedings of the IEEE conference on computer vision and pattern recognition},
  pages={499--508},
  year={2017}
}

@inproceedings{soo2017interpretable,
  title={Interpretable 3d human action analysis with temporal convolutional networks},
  author={Soo Kim, Tae and Reiter, Austin},
  booktitle={Proceedings of the IEEE conference on computer vision and pattern recognition workshops},
  pages={20--28},
  year={2017}
}

@inproceedings{yan2018spatial,
  title={Spatial temporal graph convolutional networks for skeleton-based action recognition},
  author={Yan, Sijie and Xiong, Yuanjun and Lin, Dahua},
  booktitle={Proceedings of the AAAI conference on artificial intelligence},
  volume={32},
  number={1},
  year={2018}
}

@inproceedings{shi2019two,
  title={Two-stream adaptive graph convolutional networks for skeleton-based action recognition},
  author={Shi, Lei and Zhang, Yifan and Cheng, Jian and Lu, Hanqing},
  booktitle={Proceedings of the IEEE/CVF conference on computer vision and pattern recognition},
  pages={12026--12035},
  year={2019}
}

@inproceedings{liu2020disentangling,
  title={Disentangling and unifying graph convolutions for skeleton-based action recognition},
  author={Liu, Ziyu and Zhang, Hongwen and Chen, Zhenghao and Wang, Zhiyong and Ouyang, Wanli},
  booktitle={Proceedings of the IEEE/CVF conference on computer vision and pattern recognition},
  pages={143--152},
  year={2020}
}

@inproceedings{chen2021channel,
  title={Channel-wise topology refinement graph convolution for skeleton-based action recognition},
  author={Chen, Yuxin and Zhang, Ziqi and Yuan, Chunfeng and Li, Bing and Deng, Ying and Hu, Weiming},
  booktitle={Proceedings of the IEEE/CVF international conference on computer vision},
  pages={13359--13368},
  year={2021}
}

@inproceedings{zheng2018unsupervised,
  title={Unsupervised representation learning with long-term dynamics for skeleton based action recognition},
  author={Zheng, Nenggan and Wen, Jun and Liu, Risheng and Long, Liangqu and Dai, Jianhua and Gong, Zhefeng},
  booktitle={Proceedings of the AAAI Conference on Artificial Intelligence},
  volume={32},
  number={1},
  year={2018}
}

@inproceedings{kundu2019unsupervised,
  title={Unsupervised feature learning of human actions as trajectories in pose embedding manifold},
  author={Kundu, Jogendra Nath and Gor, Maharshi and Uppala, Phani Krishna and Radhakrishnan, Venkatesh Babu},
  booktitle={2019 IEEE winter conference on applications of computer vision (WACV)},
  pages={1459--1467},
  year={2019},
  organization={IEEE}
}

@inproceedings{su2020predict,
  title={Predict \& cluster: Unsupervised skeleton based action recognition},
  author={Su, Kun and Liu, Xiulong and Shlizerman, Eli},
  booktitle={Proceedings of the IEEE/CVF conference on computer vision and pattern recognition},
  pages={9631--9640},
  year={2020}
}

@article{rao2021augmented,
  title={Augmented skeleton based contrastive action learning with momentum lstm for unsupervised action recognition},
  author={Rao, Haocong and Xu, Shihao and Hu, Xiping and Cheng, Jun and Hu, Bin},
  journal={Information Sciences},
  volume={569},
  pages={90--109},
  year={2021},
  publisher={Elsevier}
}

@inproceedings{li20213d,
  title={3d human action representation learning via cross-view consistency pursuit},
  author={Li, Linguo and Wang, Minsi and Ni, Bingbing and Wang, Hang and Yang, Jiancheng and Zhang, Wenjun},
  booktitle={Proceedings of the IEEE/CVF conference on computer vision and pattern recognition},
  pages={4741--4750},
  year={2021}
}

@inproceedings{guo2022contrastive,
  title={Contrastive learning from extremely augmented skeleton sequences for self-supervised action recognition},
  author={Guo, Tianyu and Liu, Hong and Chen, Zhan and Liu, Mengyuan and Wang, Tao and Ding, Runwei},
  booktitle={Proceedings of the AAAI Conference on Artificial Intelligence},
  volume={36},
  number={1},
  pages={762--770},
  year={2022}
}

@inproceedings{lin2023actionlet,
  title={Actionlet-dependent contrastive learning for unsupervised skeleton-based action recognition},
  author={Lin, Lilang and Zhang, Jiahang and Liu, Jiaying},
  booktitle={Proceedings of the IEEE/CVF Conference on Computer Vision and Pattern Recognition},
  pages={2363--2372},
  year={2023}
}

@article{madry2017towards,
  title={Towards deep learning models resistant to adversarial attacks},
  author={Madry, Aleksander and Makelov, Aleksandar and Schmidt, Ludwig and Tsipras, Dimitris and Vladu, Adrian},
  journal={arXiv preprint arXiv:1706.06083},
  year={2017}
}

@inproceedings{cohen2019certified,
  title={Certified adversarial robustness via randomized smoothing},
  author={Cohen, Jeremy and Rosenfeld, Elan and Kolter, Zico},
  booktitle={international conference on machine learning},
  pages={1310--1320},
  year={2019},
  organization={PMLR}
}

@inproceedings{cisse2017parseval,
  title={Parseval networks: Improving robustness to adversarial examples},
  author={Cisse, Moustapha and Bojanowski, Piotr and Grave, Edouard and Dauphin, Yann and Usunier, Nicolas},
  booktitle={International conference on machine learning},
  pages={854--863},
  year={2017},
  organization={PMLR}
}

@inproceedings{xie2019feature,
  title={Feature denoising for improving adversarial robustness},
  author={Xie, Cihang and Wu, Yuxin and Maaten, Laurens van der and Yuille, Alan L and He, Kaiming},
  booktitle={Proceedings of the IEEE/CVF conference on computer vision and pattern recognition},
  pages={501--509},
  year={2019}
}

@inproceedings{zhang2019theoretically,
  title={Theoretically principled trade-off between robustness and accuracy},
  author={Zhang, Hongyang and Yu, Yaodong and Jiao, Jiantao and Xing, Eric and El Ghaoui, Laurent and Jordan, Michael},
  booktitle={International conference on machine learning},
  pages={7472--7482},
  year={2019},
  organization={PMLR}
}

@article{kim2020adversarial,
  title={Adversarial self-supervised contrastive learning},
  author={Kim, Minseon and Tack, Jihoon and Hwang, Sung Ju},
  journal={Advances in neural information processing systems},
  volume={33},
  pages={2983--2994},
  year={2020}
}

@article{fan2021does,
  title={When does contrastive learning preserve adversarial robustness from pretraining to finetuning?},
  author={Fan, Lijie and Liu, Sijia and Chen, Pin-Yu and Zhang, Gaoyuan and Gan, Chuang},
  journal={Advances in neural information processing systems},
  volume={34},
  pages={21480--21492},
  year={2021}
}

@inproceedings{wang2023defending,
  title={Defending black-box skeleton-based human activity classifiers},
  author={Wang, He and Diao, Yunfeng and Tan, Zichang and Guo, Guodong},
  booktitle={Proceedings of the AAAI Conference on Artificial Intelligence},
  volume={37},
  number={2},
  pages={2546--2554},
  year={2023}
}

@article{van2017neural,
  title={Neural discrete representation learning},
  author={Van Den Oord, Aaron and Vinyals, Oriol and others},
  journal={Advances in neural information processing systems},
  volume={30},
  year={2017}
}

@article{ellis2013exploring,
  title={Exploring the trade-off between accuracy and observational latency in action recognition},
  author={Ellis, Chris and Masood, Syed Zain and Tappen, Marshall F and LaViola, Joseph J and Sukthankar, Rahul},
  journal={International Journal of Computer Vision},
  volume={101},
  pages={420--436},
  year={2013},
  publisher={Springer}
}

@incollection{muller2005efficient,
  title={Efficient content-based retrieval of motion capture data},
  author={M{\"u}ller, Meinard and R{\"o}der, Tido and Clausen, Michael},
  booktitle={ACM SIGGRAPH 2005 Papers},
  pages={677--685},
  year={2005}
}

@inproceedings{du2015skeleton,
  title={Skeleton based action recognition with convolutional neural network},
  author={Du, Yong and Fu, Yun and Wang, Liang},
  booktitle={2015 3rd IAPR Asian conference on pattern recognition (ACPR)},
  pages={579--583},
  year={2015},
  organization={IEEE}
}

@article{ji20123d,
  title={3D convolutional neural networks for human action recognition},
  author={Ji, Shuiwang and Xu, Wei and Yang, Ming and Yu, Kai},
  journal={IEEE transactions on pattern analysis and machine intelligence},
  volume={35},
  number={1},
  pages={221--231},
  year={2012},
  publisher={IEEE}
}

@inproceedings{he2020momentum,
  title={Momentum contrast for unsupervised visual representation learning},
  author={He, Kaiming and Fan, Haoqi and Wu, Yuxin and Xie, Saining and Girshick, Ross},
  booktitle={Proceedings of the IEEE/CVF conference on computer vision and pattern recognition},
  pages={9729--9738},
  year={2020}
}

@article{bengio2013estimating,
  title={Estimating or propagating gradients through stochastic neurons for conditional computation},
  author={Bengio, Yoshua and L{\'e}onard, Nicholas and Courville, Aaron},
  journal={arXiv preprint arXiv:1308.3432},
  year={2013}
}

@inproceedings{shahroudy2016ntu,
  title={Ntu rgb+ d: A large scale dataset for 3d human activity analysis},
  author={Shahroudy, Amir and Liu, Jun and Ng, Tian-Tsong and Wang, Gang},
  booktitle={Proceedings of the IEEE conference on computer vision and pattern recognition},
  pages={1010--1019},
  year={2016}
}

@article{liu2019ntu,
  title={Ntu rgb+ d 120: A large-scale benchmark for 3d human activity understanding},
  author={Liu, Jun and Shahroudy, Amir and Perez, Mauricio and Wang, Gang and Duan, Ling-Yu and Kot, Alex C},
  journal={IEEE transactions on pattern analysis and machine intelligence},
  volume={42},
  number={10},
  pages={2684--2701},
  year={2019},
  publisher={IEEE}
}

@article{liu2020benchmark,
  title={A benchmark dataset and comparison study for multi-modal human action analytics},
  author={Liu, Jiaying and Song, Sijie and Liu, Chunhui and Li, Yanghao and Hu, Yueyu},
  journal={ACM Transactions on Multimedia Computing, Communications, and Applications (TOMM)},
  volume={16},
  number={2},
  pages={1--24},
  year={2020},
  publisher={ACM New York, NY, USA}
}

@inproceedings{athalye2018obfuscated,
  title={Obfuscated gradients give a false sense of security: Circumventing defenses to adversarial examples},
  author={Athalye, Anish and Carlini, Nicholas and Wagner, David},
  booktitle={International conference on machine learning},
  pages={274--283},
  year={2018},
  organization={PMLR}
}

@inproceedings{wang2021understanding,
  title={Understanding the robustness of skeleton-based action recognition under adversarial attack},
  author={Wang, He and He, Feixiang and Peng, Zhexi and Shao, Tianjia and Yang, Yong-Liang and Zhou, Kun and Hogg, David},
  booktitle={Proceedings of the IEEE/CVF Conference on Computer Vision and Pattern Recognition},
  pages={14656--14665},
  year={2021}
}

@inproceedings{croce2020reliable,
  title={Reliable evaluation of adversarial robustness with an ensemble of diverse parameter-free attacks},
  author={Croce, Francesco and Hein, Matthias},
  booktitle={International conference on machine learning},
  pages={2206--2216},
  year={2020},
  organization={PMLR}
}

@inproceedings{diao2021basar,
  title={BASAR: Black-box attack on skeletal action recognition},
  author={Diao, Yunfeng and Shao, Tianjia and Yang, Yong-Liang and Zhou, Kun and Wang, He},
  booktitle={Proceedings of the IEEE/CVF conference on computer vision and pattern recognition},
  pages={7597--7607},
  year={2021}
}
\end{document}